\documentclass[conference]{IEEEtran}
\usepackage{graphics}      
\usepackage[T1]{fontenc}   
\usepackage{txfonts}
\usepackage{mathptmx}
\usepackage[pdflang={en-US},pdftex]{hyperref}
\usepackage{color}
\usepackage{booktabs}
\usepackage{textcomp}
\usepackage{float}
\usepackage{enumitem}
\usepackage{cite}
\usepackage{amssymb,amsfonts}
\usepackage{algorithmic}
\usepackage{graphicx}
\usepackage{textcomp}
\usepackage{xcolor}
\usepackage{booktabs}
\usepackage{tabularx}
\usepackage{algorithm}
\usepackage{algorithmic}
\usepackage{outlines}

\def\BibTeX{{\rm B\kern-.05em{\sc i\kern-.025em b}\kern-.08em
    T\kern-.1667em\lower.7ex\hbox{E}\kern-.125emX}}

\DeclareRobustCommand*{\IEEEauthorrefmark}[1]{%
\raisebox{0pt}[0pt][0pt]{\textsuperscript{\footnotesize #1}}%
}

\begin{document}

\title{Towards Evaluating Gaussian Blurring in Perceptual Hashing as a Facial Image Filter \\
}

\author{\IEEEauthorblockN{Yigit Alparslan\IEEEauthorrefmark{1} Ken Alparslan\IEEEauthorrefmark{2} 
Mannika Kshettry\IEEEauthorrefmark{3} Dr. Louis Kratz\IEEEauthorrefmark{4}}
\IEEEauthorblockA{\IEEEauthorrefmark{1,}\IEEEauthorrefmark{4}Department of Computer Science, Drexel University, Philadelphia, PA, US 
\\
\IEEEauthorrefmark{2}Department of Computer Science, Conestoga College, Waterloo, ON, CA\\
\IEEEauthorrefmark{3}Electrical and Computer Engineering Drexel University, Philadelphia, PA, US\\
Email: \{ya332\IEEEauthorrefmark{1}, mk3442\IEEEauthorrefmark{3}, lak24\IEEEauthorrefmark{3}\}@drexel.edu, kalparslan6724@conestogac.on.ca\IEEEauthorrefmark{2} }}

\maketitle

\begin{abstract}
With the growth in social media, there is a huge amount of images of faces available on the internet. Often, people use other people's pictures on their own profile. Perceptual hashing is often used to detect whether two images are identical. Therefore, it can be used to detect whether people are misusing others’ pictures. In perceptual hashing, a hash is calculated for a given image, and a new test image is mapped to one of the existing hashes if duplicate features are present. Therefore, it can be used as an image filter to flag banned image content or adversarial attacks -- which are modifications that are made on purpose to deceive the filter-- even though the content might be changed to deceive the filters. For this reason, it is critical for perceptual hashing to be robust enough to take transformations such as resizing, cropping, and slight pixel modifications into account. In this paper, we would like to propose to experiment with effect of Gaussian blurring in perceptual hashing for detecting misuse of personal images specifically for face images. We hypothesize that use of Gaussian blurring on the image before calculating its hash will increase the accuracy of our filter that detects adversarial attacks which consist of image cropping, adding text annotation, and image rotation.\end{abstract}


\section{Introduction}

Many websites that support image uploading, media sharing, and social profile creation have filters to detect banned image content. Such filters are crucial to creating a safe and functional media sites. With recent breakthroughs in machine learning and deep learning, a new set of perturbations called adversarial attacks have been tested several times to result in misclassifications by neural networks. Convolutional neural networks have been used in several domains such as surveillance, spam detection, autonomous driving, crime fighting, malware detection [12]. Therefore, it is very crucial that neural networks that are deployed in such mission critical systems are robust enough to such adversarial attacks. Often, adversarials try to deceive such filters and convolutional neural networks by incorporating slight modifications to images [9][10][11]. Such adversarial attacks can be image annotation, resizing, cropping etc. Such filters need to be robust enough to flag the banned image content even though there are a small amount of perturbations. Besides banned content, filters are also used for verifying image content, image authentication and watermark. Such filters can be built via perceptual hashing. Perceptual hashing is simple, fast, yet powerful[1]. It is based on the idea of creating similar hashes for similar content[2]. If a new image’s hash is very similar to an existing copyrighted image, then the new image can be flagged as copyright violation.
 
Currently, effect of perceptual hashing was proven to work with images that are JPEG compressed, Gaussian blurred (after hash was computed), or noised [8]. In such studies, Gaussian smoothing was applied as an adversarial attack to the test images. We would like to hypothesize that applying Gaussian smoothing to the input images before calculating the hash distances for our baseline images will increase the filter accuracy. We use Yale B Extended Face Data set to see the effect of Gaussian smoothing on facial image filters. The adversarial attacks in this paper are text annotation, rotation by 180 degrees, rotation by 45 degrees, and cropping around 15\% of the original image.

\section{Related Work}
Numerous studies showed that perceptual hashing is a powerful and simple algorithm for image content verification, multimedia watermarking, and image authentication[3][4][5][6][7] Currently there exists an algorithm that computes a robust intermediate hash for images when several adversarial attacks and perturbations are added to the image[8]. The perceptual hashing in this paper is tolerant to JPEG compression, resizing, and Gaussian Blurring. However, the paper is using Gaussian blurring as an adversarial attack after the hash values are calculated, and the data set is not a facial dataset. We believe there is value in studying the perceptual hashing with face images because of the popularity of face detection technologies. Our novelty is to use a facial dataset to deter misuse of public profile pictures such as unauthorized image sharing in other website. We also would like to use Gaussian blurring before calculating the hash functions to leverage the image content, and see if there is any effect on the accuracy in our adversarial detection. Also, another novelty that we bring with our paper is investigating the addition of text annotation on perceptual hashing, which was not implemented on the paper.

\section{Method}
Our general approach can be quickly summarized as follows:
\begin{outline}[enumerate]
    \1 Use Yale Face Database B dataset for our experiments(576 poses per 28 human subjects)
    \1 Calculate hash values for about 28 images(1 of each subject) in the dataset as a baseline, where we first
    \2	Calculate DCT frequency coefficients
    \2	Get top 64 coefficients to calculate an 64 bit hash code( We anticipate 64 coefficients will be enough, we might need to experiment with this number, i.e. 128 or 32)(Note at this step we likely will need to resize/sample the images)
    \1 Save those 28 hash codes as baseline images.
    \1 Repeat the previous hash computation but make sure to blur the image with Gaussian Blurring before calculating the coefficients(sigma and kernel size need to be big enough so that image seems blurred to human eye)
    \1 Save those 28 hash codes as Gaussian Blurred ‘GB’ images.
    \1 Now create the malicious content/adversarial attacks by doing the following to all 28 images.
	\2  Cropping (so that hair and neck does not show up and just face appears)
    \2	Adding text on images (a simple text -unfortunately it is too early to determine the font size, font type or color, but we anticipate to add something simple such as a black text across the image saying ‘copyrighted’. Since we are aiming for robust face image detection, font size or font color should not matter on theory)
    \2	Rotating images 180 degrees(so upside down)
    \2 Rotating images 45 degrees(Crop the black pixels at the sides if the image is no longer rectangular)
    \1 Use the above altered images as the test dataset
    \1 Compute hash values for the test dataset.
    \1 Test if the test dataset images are detected as duplicates of the baseline images or as duplicates of the ‘GB’ images.
    \1 Verify if the hypothesis failed/succeeded.
    \1 Conclude with an explanation of the results, and determine whether future work is necessary.
\end{outline}

Having outlined the requirements, now we expand upon them in detail here.

\subsection{Data}
Perceptual hashing was used for facial detection and to detect malicious use of other’s images. We used 28 images from the Yale B Extended Face Dataset for creating our dataset for this experiment. Each image was blurred using the Gaussian Blurring technique. A Baseline dataset was created for with all the original images and a Blurred dataset was created with all the Gaussian Blurred images. We are open-sourcing our code\footnote{https://github.com/ya332/Perceptual-Hashing-as-Adversarial-Defense} to further foster improvements about our study in scientific community. (Our source code also has all the requirements outlined in the README section)

\subsection{Adversarial Attacks}
For our experiments we created a test dataset by manipulating each of the images. Our manipulations included - (i) adding text annotations on the image, (ii) cropping the images from all sides, (iii) rotating the image by 180 degrees, and (iv) rotating the image by 45 degrees. Fig. 1 shows an example of our test images.

\begin{figure}[!ht]
\centering
  \includegraphics[width=0.9\columnwidth]{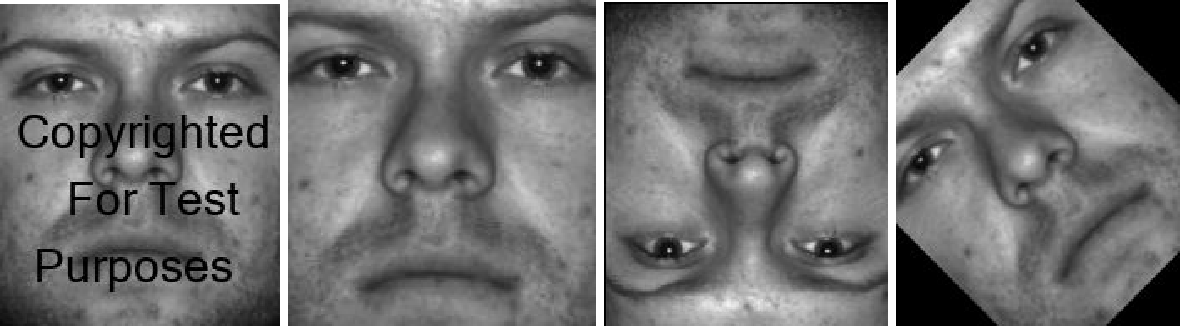}
  \caption{Test dataset example. (a) Annotated Image (b) Cropped Image (c) 180 Degrees Rotated Image (d) 45 degrees Rotated Image}~\label{fig:figure1}
\end{figure}

\section{Experiment Results and Discussion}

We tested different threshold values for each of our test subsets - (a) Annotated, (b) Cropped Image, (c) 180 Degrees Rotated Image, and (d) 45 degrees Rotated Image. We varied the threshold for both Average hash and Discrete Cosine Transform hash and for both 32-bit and 64-bit hashes.

\textbf{Quick Summary of Our Results}
\begin{enumerate}
\item	Our hypothesis was that we could improve the accuracy if we blurred the image before calculating its hash. However, there was no significant improvement seen in the accuracy when the Blurred Images were used.

\item	For Average Hash, the 64-bit hash value gave better results with smaller threshold values in the Hamming Distance function.

\item	The 32-bit DCT Hash gave the best results for annotated images.

\item	The accuracy only improved for the cropped images. However, the increase was only by 3.6\%.

\item	The Discrete Cosine Transform (DCT) hash worked better than the Average hash for the annotated images.

\item	In the case of the cropped dataset, both the DCT and Average hash had an equal accuracy for 64-Bit hashes.

\item	Rotation by 45 degrees had 0\% accuracy(regardless of hash function type or hash length)

\end{enumerate}

Having outlined the results, now we expand upon them in detail here.
\subsection{Annotated Images}Our algorithm worked best for the annotated images. We were able to achieve around 80\% accuracy. The accuracy for 64-bit Average hash was the same for the Baseline and the Blurred images (Fig. 2). At a threshold value of 16 we were able to achieve an accuracy of 78.5\%. The accuracy for 32-bit Average hash was the same for the Baseline and the Blurred images (Fig.

3). At a threshold value of 19 we got an accuracy 75.0\%. For the 64-bit Discrete Cosine Transform (DCT) Hash we got an accuracy of 82.1\% for the threshold value of 16 for both Baseline and Blurred Images (Fig. 4). We got an 85\% accuracy with threshold value 6 for the
32-bit DCT Hash (Fig. 5). Our hypothesis was that we could improve the accuracy if we blurred the image before calculating its hash. There was no improvement seen in the accuracy when the Blurred Images were used. All the plots for the annotated images have a spike, exponential growth and a plateau after a certain threshold value in their profile. The number of bits changed is high enough that all the hashes of the annotated images map to their respective original images. This gives a 100\% accuracy, but it also wrongly maps other images which does not make it the optimal threshold value. For Average Hash, the 64-bit hash value gave better results with lower number of bit changes. The 32-bit DCT Hash gave the best results for annotated images.

\begin{figure}[!ht]
\centering
  \includegraphics[width=0.9\columnwidth]{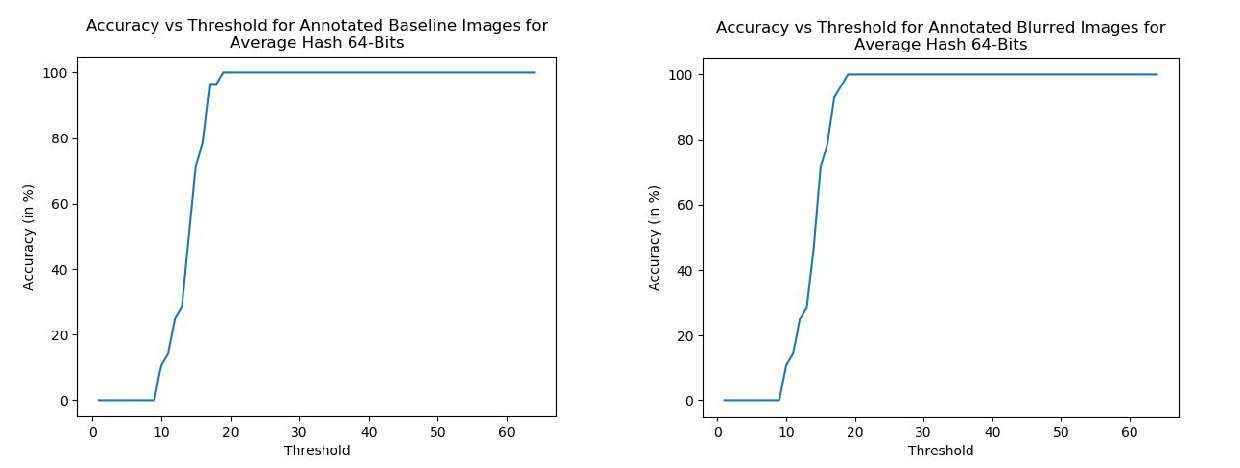}
  \caption{The plots for Accuracy vs Threshold are shown. These plots are for 64-Bit Average Hash for Annotated Images. (a) Plot for Baseline Images (b) Plot for Blurred Images}~\label{fig:figure2}
\end{figure}

\begin{figure}[!ht]
\centering
  \includegraphics[width=0.9\columnwidth]{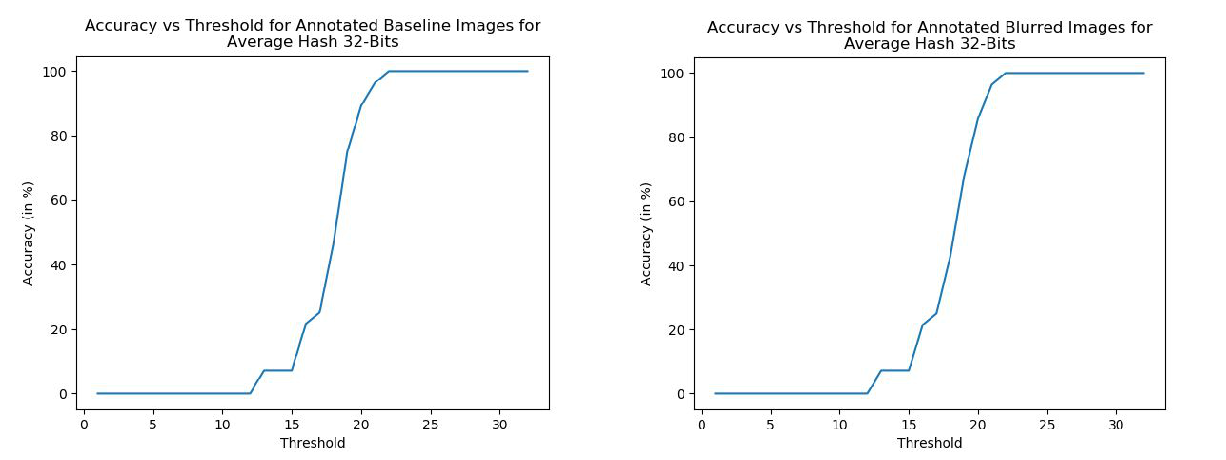}
  \caption{The plots for Accuracy vs Threshold are shown. These plots are for 32-Bit Average Hash for Annotated Images. (a) Plot for Baseline Images (b) Plot for Blurred Images}~\label{fig:figure3}
\end{figure}

 \begin{figure}[!ht]
\centering
  \includegraphics[width=0.9\columnwidth]{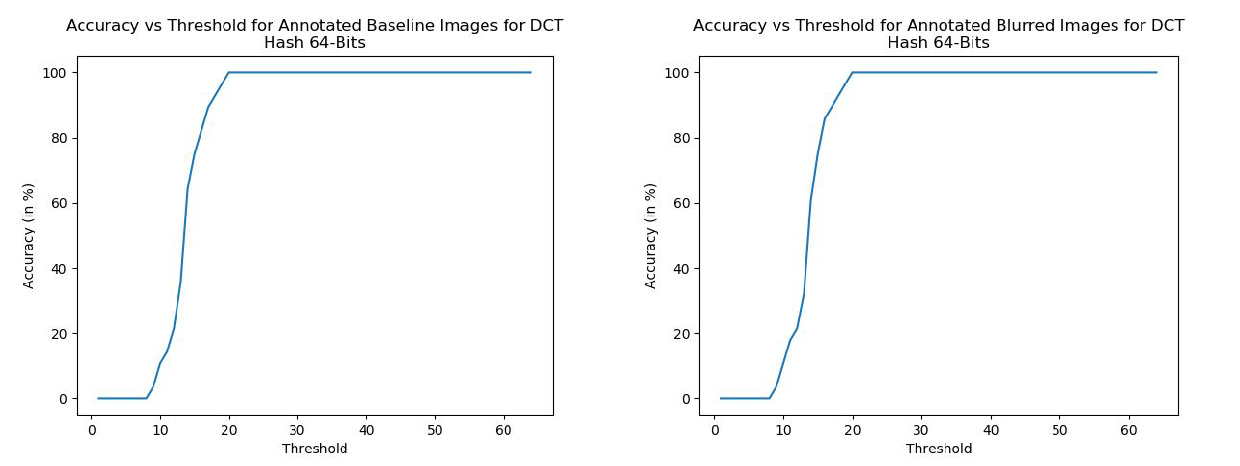}
  \caption{The plots for Accuracy vs Threshold are shown. These plots are for 64-Bit DCT Hash for Annotated Images. (a) Plot for Baseline Images (b) Plot for Blurred Images}~\label{fig:figure4}
\end{figure}

 \begin{figure}[!ht]
\centering
  \includegraphics[width=0.9\columnwidth]{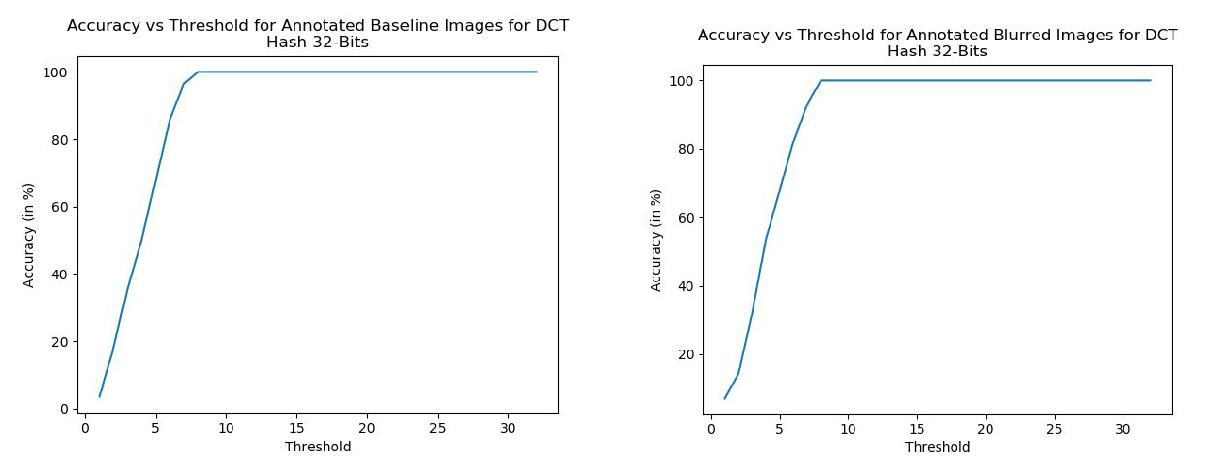}
  \caption{The plots for Accuracy vs Threshold are shown. These plots are for 32-Bit DCT Hash for Annotated Images. (a) Plot for Baseline Images (b) Plot for Blurred Images
}~\label{fig:figure5}
\end{figure}

\subsection{Cropped Images}

The accuracy for finding cropped duplicates was very low. The accuracy for Baseline images was 14.3\% and for Blurred images it was 17.9\%. This was for 64-bit Average hash and a threshold value of 14 (Fig. 6). The accuracy improved with a 32-bit Average Hash. It was 46.4\% for the Baseline Images and 50\% for Blurred Images (Fig. 7). But this accuracy was achieved for a threshold value of 18. Using this high threshold value will also result in some false mappings. The accuracy for 64-bit DCT has was 17.8\% for Baseline images and 21.4\% for Blurred images. The threshold value for this accuracy 14 (Fig. 8). The accuracy for 32-bit DCT Hash was almost nil (Fig. 9).

\begin{figure}[!ht]
\centering
  \includegraphics[width=0.9\columnwidth]{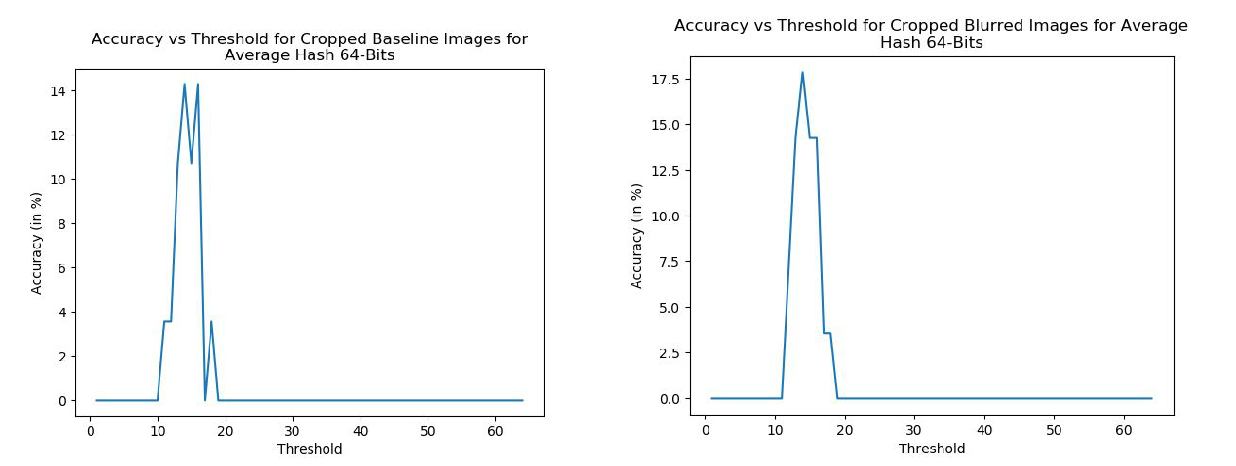}
  \caption{The plots for Accuracy vs Threshold are shown. These plots are for 64-Bit Average Hash for Cropped Images. (a) Plot for Baseline Images (b) Plot for Blurred Images}~\label{fig:figure6}
\end{figure}

\begin{figure}[!ht]
\centering
  \includegraphics[width=0.9\columnwidth]{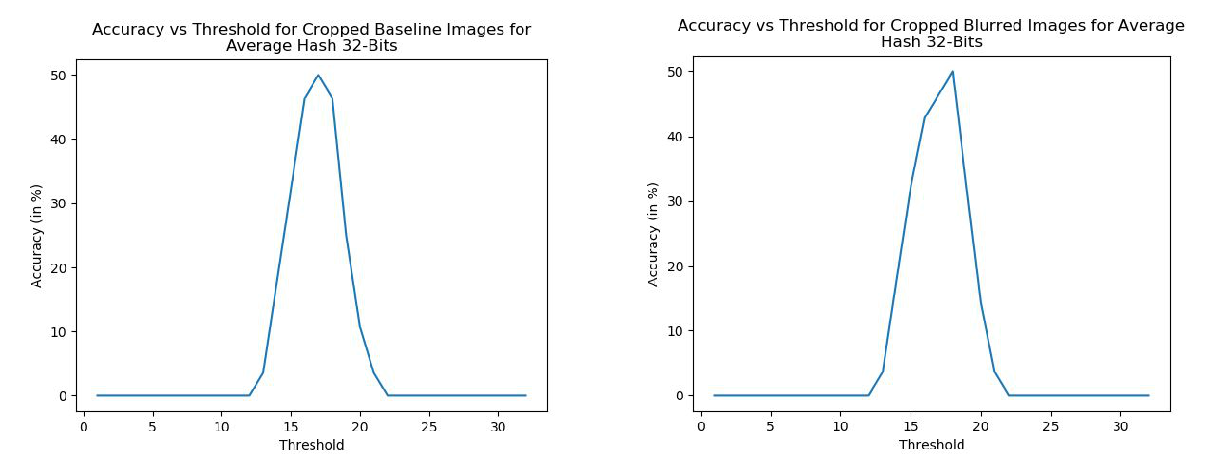}
  \caption{The plots for Accuracy vs Threshold are shown. These plots are for 32-Bit Average Hash for Cropped Images. (a) Plot for Baseline Images (b) Plot for Blurred Images}~\label{fig:figure7}
\end{figure}

\begin{figure}[!ht]
\centering
  \includegraphics[width=0.9\columnwidth]{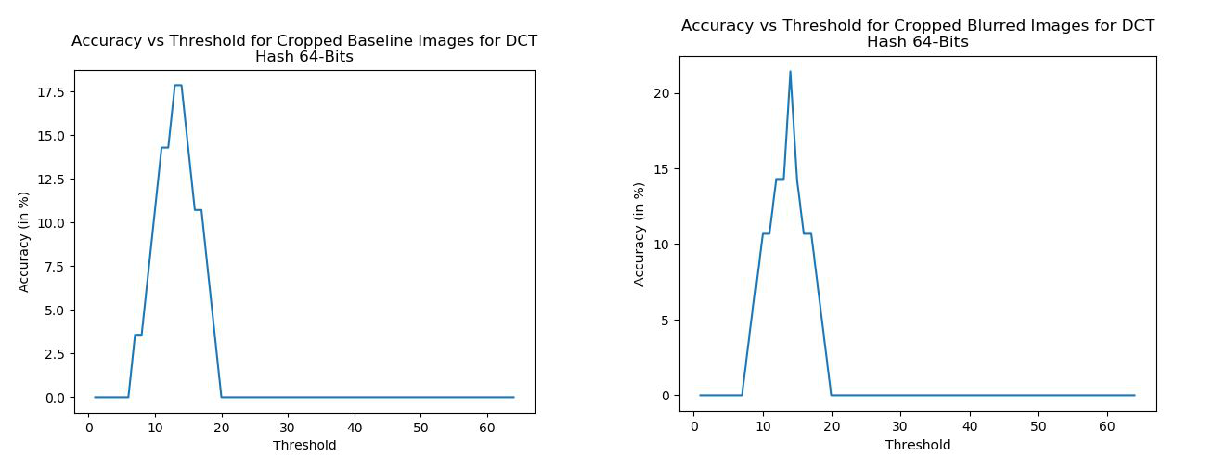}
  \caption{The plots for Accuracy vs Threshold are shown. These plots are for 64-Bit DCT Hash for Cropped Images. (a) Plot for Baseline Images (b) Plot for Blurred Images}~\label{fig:figure8}
\end{figure}

\begin{figure}[!ht]
\centering
  \includegraphics[width=0.9\columnwidth]{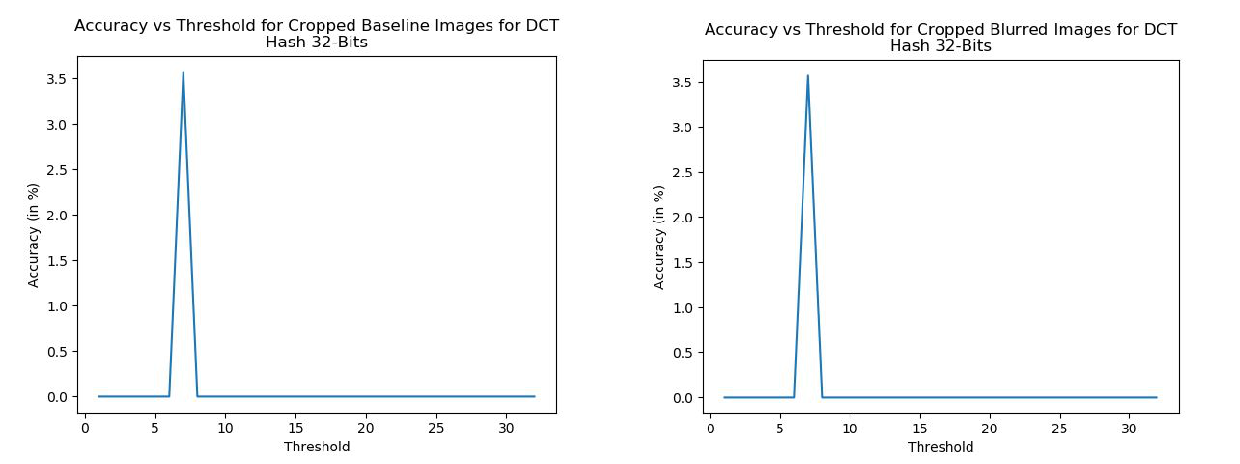}
  \caption{The perceptual hashing algorithm does not work for rotated images. We were able to achieve a 25-30\% accuracy for 64-Bit Average hash for 180 degrees rotated image (Fig. 10). The accuracy was lower for the 32-bit average hash (Fig. 11). The accuracy was low for both 64-Bit and 32-Bit Average hashes for the 45 degrees rotated images (Fig. 14, Fig.15). The accuracy was almost nil for 64-Bit and 32-Bit DCT Hash for both 180 degrees rotated and 45 degrees rotated images. (Fig. 12, Fig. 13, Fig. 16, Fig. 17)}~\label{fig:figure9}
\end{figure}

\subsection{Rotated Images}

The perceptual hashing algorithm does not work for rotated images. We were able to achieve a 25-30\% accuracy for 64-Bit Average hash for 180 degrees rotated image (Fig. 10). The accuracy was lower for the 32-bit average hash (Fig. 11). The accuracy was low for both 64-Bit and 32-Bit Average hashes for the 45 degrees rotated images (Fig. 14, Fig.15). The accuracy was almost nil for 64-Bit and 32-Bit DCT Hash for both 180 degrees rotated and 45 degrees rotated images. (Fig. 12, Fig. 13, Fig. 16, Fig. 17)

\begin{figure}[!ht]
\centering
  \includegraphics[width=0.9\columnwidth]{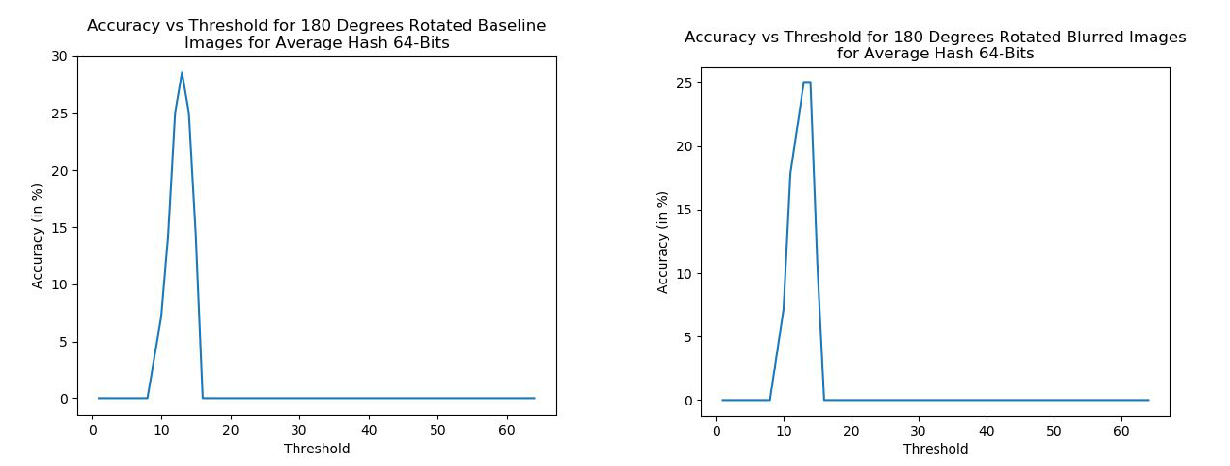}
  \caption{The plots for Accuracy vs Threshold are shown. These plots are for 64-Bit Average Hash for 180 Degrees Rotated Images. (a) Plot for Baseline Images (b) Plot for Blurred Images
 }~\label{fig:figure10}
\end{figure}

\begin{figure}[!ht]
\centering
  \includegraphics[width=0.9\columnwidth]{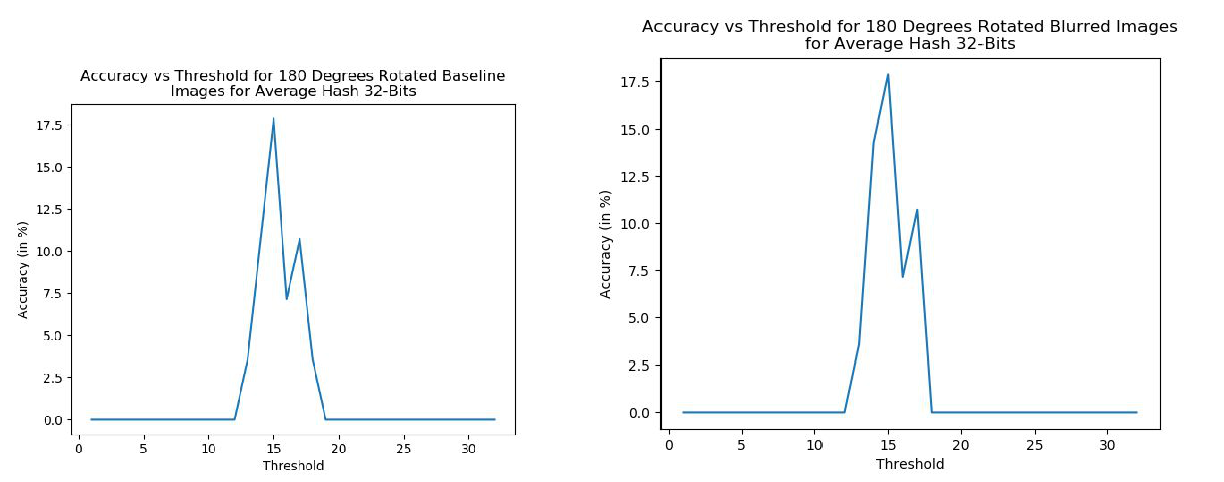}
  \caption{The plots for Accuracy vs Threshold are shown. These plots are for 32-Bit Average Hash for 180 Degrees Rotated Images. (a) Plot for Baseline Images (b) Plot for Blurred Images}~\label{fig:figure11}
\end{figure}

\begin{figure}[!ht]
\centering
  \includegraphics[width=0.9\columnwidth]{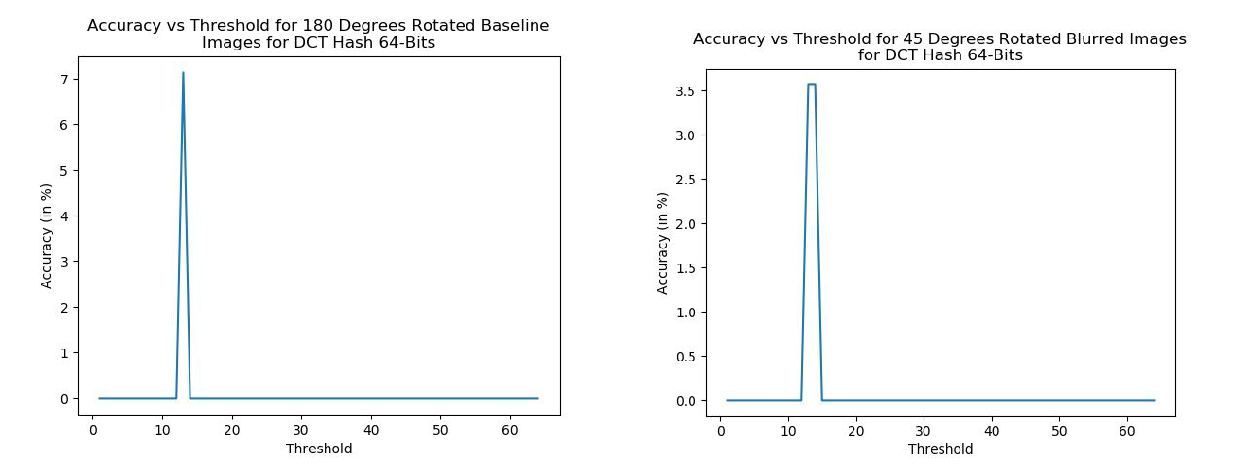}
  \caption{The plots for Accuracy vs Threshold are shown. These plots are for 64-Bit DCT Hash for 180 Degrees Rotated Images. (a) Plot for Baseline Images (b) Plot for Blurred Images
 }~\label{fig:figure12}
\end{figure}

\begin{figure}[!ht]
\centering
  \includegraphics[width=0.9\columnwidth]{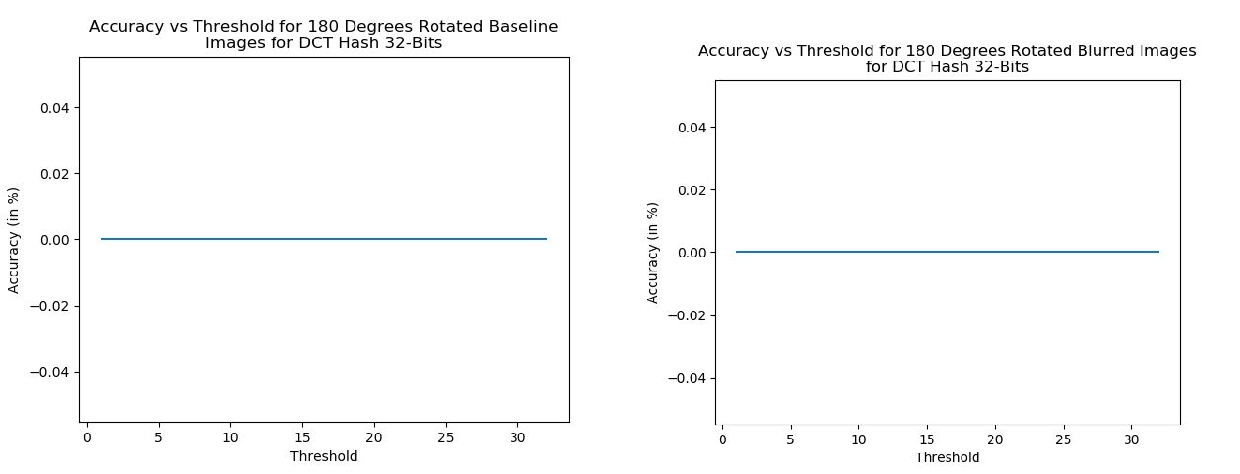}
  \caption{The plots for Accuracy vs Threshold are shown. These plots are for 64-Bit DCT Hash for 180 Degrees Rotated Images. (a) Plot for Baseline Images (b) Plot for Blurred Images
 }~\label{fig:figure13}
\end{figure}

\begin{figure}[!ht]
\centering
  \includegraphics[width=0.9\columnwidth]{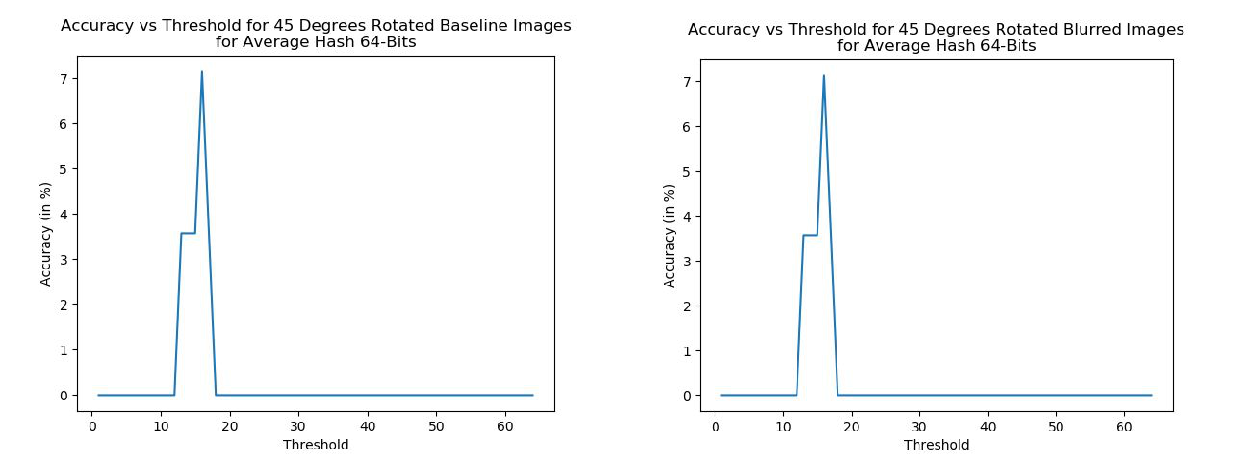}
  \caption{The plots for Accuracy vs Threshold are shown. These plots are for 64-Bit DCT Hash for 180 Degrees Rotated Images. (a) Plot for Baseline Images (b) Plot for Blurred Images
 }~\label{fig:figure14}
\end{figure}

\begin{figure}[!ht]
\centering
  \includegraphics[width=0.9\columnwidth]{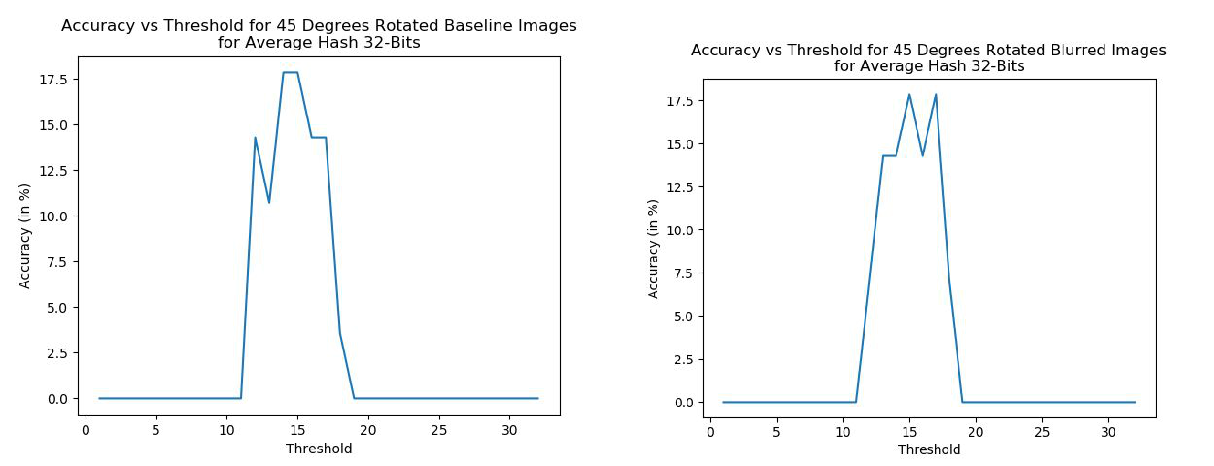}
  \caption{The plots for Accuracy vs Threshold are shown. These plots are for 64-Bit DCT Hash for 180 Degrees Rotated Images. (a) Plot for Baseline Images (b) Plot for Blurred Images
 }~\label{fig:figure15}
\end{figure}

\begin{figure}[!ht]
\centering
  \includegraphics[width=0.9\columnwidth]{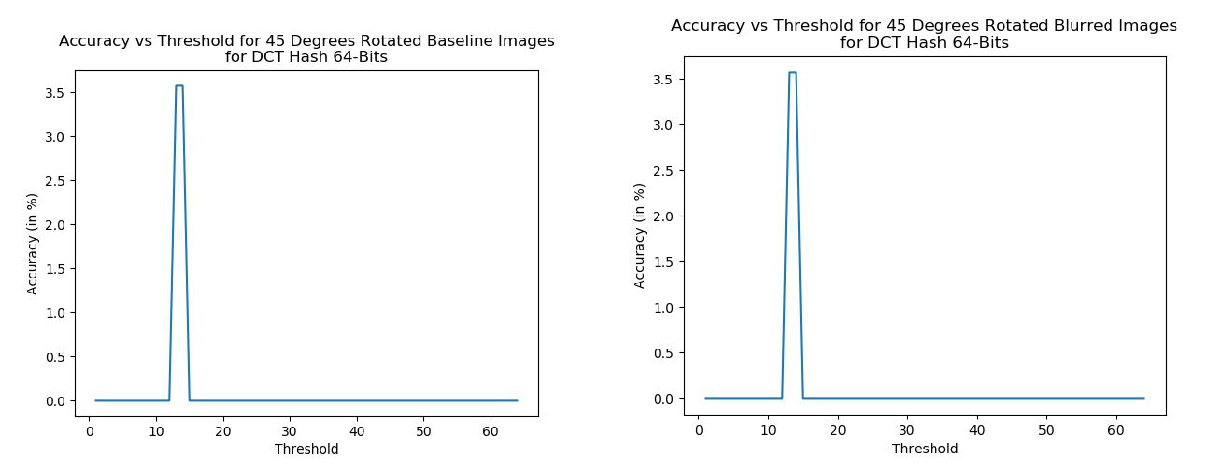}
  \caption{The plots for Accuracy vs Threshold are shown. These plots are for 64-Bit DCT Hash for 180 Degrees Rotated Images. (a) Plot for Baseline Images (b) Plot for Blurred Images
 }~\label{fig:figure16}
\end{figure}

\begin{figure}[!ht]
\centering
  \includegraphics[width=0.9\columnwidth]{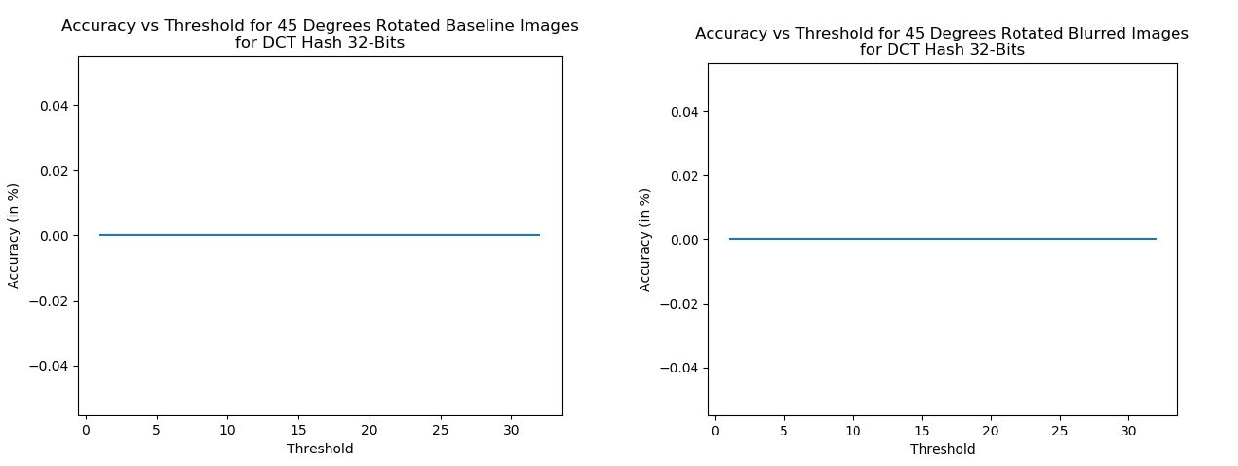}
  \caption{The plots for Accuracy vs Threshold are shown. These plots are for 64-Bit DCT Hash for 180 Degrees Rotated Images. (a) Plot for Baseline Images (b) Plot for Blurred Images
 }~\label{fig:figure17}
\end{figure}

\section{Conclusion}
Gaussian Blurring the images before computing the hash did not have a significant improvement in the accuracy of our sub-datasets. The accuracy only improved for the cropped images. However, the increase was only by 3.6\%. The Discrete Cosine Transform (DCT) hash worked better than the Average hash for the annotated images. In the case of the cropped dataset, both the DCT and Average hash had an equal accuracy for 64-Bit hashes. In most of the cases, it was seen that the 64-Bit hash worked better.

\section{Future Work}
Image dataset can be enlarged in the future. We would like to investigate the effect of colorful and grey scale image differences in the future. Additionally, we would like to investigate the 0\% accuracy on the 45 degree rotation, as well as look at other rotation angles such as 60, or 90 degrees to see if there is any significant differences.

\end{document}